# Definition sets for the Direct Kinematics of Parallel Manipulators

Philippe WENGER, Damien CHABLAT
Institut de Recherche en Cybernétique de Nantes UMR 6597 (École Centrale de Nantes / Université de Nantes)
1, rue de la Noë, BP 92101 Nantes Cedex 3 France
email : Philippe.Wenger@lan.ec-nantes.fr

**ABSTRACT**

The aim of this paper is to characterize the uniqueness domains in the workspace of parallel manipulators, as well as their image in the joint space. The notion of *aspect* introduced for serial manipulators in **[Borrel 86]** is redefined for such parallel manipulators. Then, it is shown that it is possible to link several solutions to the direct kinematic problem without meeting a singularity, thus meaning that the aspects are not uniqueness domains. Additional surfaces are characterized in the workspace which yield new uniqueness domains. An octree model of spaces is used to compute the joint space, the workspace and all other newly defined sets. This study is illustrated all along the paper with a 3-RPR planar parallel manipulator.

**Key-words** : Parallel Manipulator, Workspace, Singularity, Aspect, Uniqueness domains, Octree.

## 1. INTRODUCTION

A well known feature of parallel manipulators is the existence of multiple solutions to the direct kinematic problem. That is, the mobile platform can admit several positions and orientations (or configurations) in the workspace for one given set of input joint values **[Merlet 90]**. The dual problem arises in serial manipulators, where several input joint values correspond to one given configuration of the end-effector. To cope with the existence of multiple inverse kinematic solutions in serial manipulators, the notion of *aspects* was introduced **[Borrel 86]**. The aspects were defined as the maximal singularity-free domains in the joint space. For usual industrial serial manipulators, the aspects were found to be the maximal sets in the joint space where there is only one inverse kinematic solution. Many other serial manipulators, referred to as *cuspidal* manipulators, were shown to be able to change solution without passing through a singularity, thus meaning that there is more than one inverse kinematic solution in one aspect. New uniqueness domains have been characterized for cuspidal manipulators **[Wenger 92]**, **[El Omri 96]**. It is also of interest to be able to characterize the uniqueness domains for parallel manipulators, in order to separate and to identify, in the workspace, the different solutions to the direct kinematic problem. To the authors knowledge, the only work concerned with this issue is that of **[Chételat 96]**, which proposes a generalization of the implicit function theorem. Unfortunately, the hypothesis of convexity required by this new theorem is still too restrictive. This paper is organized as follows. Section 2 describes the planar 3-RPR parallel manipulator which will be used all along this paper to illustrate the new theoretical results. Section 3 restates the notion of *aspect* for parallel manipulators. New surfaces, the *characteristic surfaces*, are defined in section 4, which, together with the singular surfaces, further divide the aspects into smaller regions, called *basic regions*. Finally, the uniqueness domains are defined in section 5. The workspace, the aspects, the characteristic and singular surfaces, and the uniqueness domains are calculated for the planar 3-RPR parallel manipulator using octrees. The images in the joint space of the uniqueness domains are also calculated. It is shown that the joint space is composed of several subspaces with different numbers of direct kinematic solutions.

## 2. PRELIMINARIES

### 2.1 PARALLEL MANIPULATOR STUDIED

This work deals with those parallel manipulators which have only one inverse kinematic solution. In addition, the passive joints will always be assumed unlimited in this study. For more legibility, a planar manipulator will be used as illustrative example all along this paper. This is a planar 3-DOF manipulator, with 3 parallel RPR chains (Figure 1). The input joint variables are the three prismatic actuated joints. The output variables are the positions and orientation of the platform in the plane. This manipulator has been frequently studied, in particular by **[Merlet 90]**, **[Gosselin 91]** and **[Innocenti 92]**.

The kinematic equations of this manipulator are **[Gosselin 91]** :

$$\rho_1^2 = x^2 + y^2 \qquad (1)$$

$$\rho_2^2 = (x + l_2 \cos(\phi) - c_2)^2 + (y + l_2 \sin(\phi))^2 \qquad (2)$$

$$\rho_3^2 = (x + l_3 \cos(\phi + \theta) - c_3)^2 + (y + l_3 \sin(\phi + \theta) - d_3)^2 \qquad (3)$$

The dimensions of the platform are the same as in **[Merlet 90]** and in **[Innocenti 92]** :

- $A_1 = (0.0, 0.0)$    $B_1B_2 = 17.04$

- $A_2 = (15.91, 0.0)$   $B_2B_3 = 16.54$
- $A_3 = (0.0, 10.0)$   $B_3B_1 = 20.84$

The limits of the prismatic actuated joints are those chosen in **[Innocenti 92]** :

$$10.0 \leq \rho_i \leq 32.0$$

The passive revolute joints are assumed unlimited.

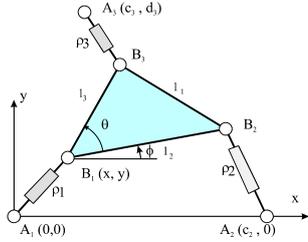

*Figure 1 : 3-RPR planar manipulator*

### 2.2 OCTREE MODEL

The octree is a hierarchical data structure based on a recursive subdivision of space **[Meagher 81]**. It is particularly useful for representing complex 3-D shapes, and is suitable for Boolean operations like union, difference and intersection. Since the octree structure has an implicit adjacency graph, arcwise-connectivity analyses can be naturally achieved. The octree model of a space S leads to a representation of S with cubes of various sizes. Basically, the smallest cubes lie near the boundary of the shape and their size determines the accuracy of the octree representation. Octrees have been used in several robotic applications **[Faverjon 84]**, **[Garcia 89]**, **[El Omri 93]**. In this work, the octree models are calculated using discretization and enrichment techniques as described in **[Chablat 96a]**.

The octree models of the joint space (in the space $\rho_1$, $\rho_2$, $\rho_3$) and of the workspace (in the space $x$, $y$ et $\phi$) of the 3-RPR parallel manipulator are shown in figures 2 and 3. The joint space is not a complete parallelepiped, since any joint vector can't lead to an assembly configuration of the manipulator.

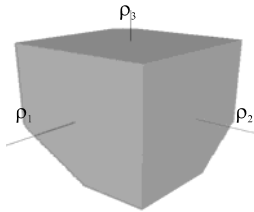 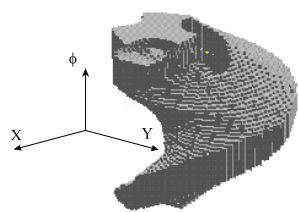

*Figure 2 : Octree model of the joint space*   *Figure 3 : Octree model of the workspace*

### 2.3 SINGULARITIES

The vector of input variables and the vector of output variables for a n-DOF parallel manipulator are related through a system of non linear algebraic equations which can be written as :

$$F(q, X) = 0 \quad (4)$$

where $0$ means here the n-dimensional zero vector.

Differentiating (4) with respect to time leads to the velocity model :

$$A\dot{X} + B\dot{q} = 0 \quad (5)$$

where $A$ and $B$ are $n \times n$ Jacobian matrices. Those matrices are functions of $q$ and $X$ :

$$A = \frac{\partial F}{\partial X} \qquad B = \frac{\partial F}{\partial q} \quad (6)$$

These matrices are useful for the determination of the singular configurations **[Sefrioui 92]**.

#### 2.3.1 Type 1 singularities

These singularities occur when $det(B) = 0$.

For the planar manipulator, this condition can be satisfied only when $\rho_1 = 0$ or $\rho_2 = 0$ or $\rho_3 = 0$.

In practise, the type-1 singularities are attained when one of the actuated prismatic joints reaches its limit [**Gosselin 90**]. The corresponding configurations are located at the boundary of the workspace.

For parallel manipulators which may have more than one inverse kinematic solutions, type-1 singularities are configurations where two solutions to the inverse kinematic problem meet. By hypothesis, type-1 singularities will be always associated with joint limits in this paper.

#### 2.3.2 Type 2 singularities

They occur when $det(A) = 0$. Unlike the preceding ones, such singular configurations occur inside the workspace. They correspond to configurations for which two branches of the direct kinematic problem meet. They are particularly undesirable since the manipulator cannot be steadily controlled in such configuration where the manipulator stiffness vanishes in some direction.

For the planar manipulator, such configurations are reached whenever the axes of the three prismatic joints intersect (possibly at infinity). In such configurations, the manipulator cannot resist a wrench applies at the intersecting point (Figure 4).

These singular configurations have been built and modelled using octrees (Figure 5). The equation of $det(A)$ can be put in an explicit form $y = s(x, \phi)$, only two variables need to be swept in the octree enrichment process **[Chablat 96b]**.

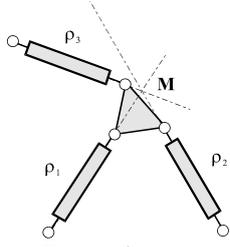
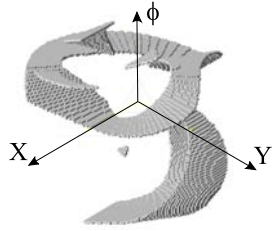

*Figure 4 :*
*Type-2 singular configurations*

*Figure 5 :*
*Octree model of the set of type-2 singularities in the workspace*

## 3. NOTION OF ASPECT FOR PARALLEL MANIPULATORS

The notion of aspect was introduced by **[Borrel 86]** to cope with the existence of multiple inverse kinematic solutions in serial manipulators. An equivalent definition was used in **[Khalil 96]** for a special case of parallel manipulators, but no formal, more general definition has been set. The aspects are redefined formally in this section.

### 3.1 DEFINITION

Let $OS_m$ and $JS_n$ denote the operational space and the joint space, respectively. $OS_m$ is the space of configurations of the moving platform and $JS_n$ is the space of the actuated joints.

Let $g$ be the map relating the actuated joint vectors with the moving platform configurations:

$$g : OS_m \to JS_n \\ X \to q = g(X) \quad (7)$$

It is assumed $m = n$, that is, only non-redundant manipulators will be studied in this paper. Finding the solutions $X$ to equation $q = g(X)$ means solving the direct kinematic problem. For our planar parallel manipulator, it has been shown that the direct kinematic model can admit six real solutions **[Gosselin 91]**.

Let $W$ be the reachable workspace, that is, the set of all positions and orientations reachable by the moving platform **[Kumar 92]**, **[Pennock 93]**. Let $Q$ be the reachable joint space that is, the set of all joint vectors reachable by actuated joints.

$$Q = \{q \in JS_n, \forall i \leq n, q_{i\,min} \leq q_i \leq q_{i\,max}\}, Q \subset JS_n \quad (8)$$

$$Q = g(W), W \subset OS_m \quad (9)$$

### Definition 1:

The aspects $WA_i$ is defined as the maximal sets such that:

- $WA_i \subset W$;
- $WA_i$ is connected;
- $\forall X \in WA_i$, $Det(WA_i) \neq 0$.

In other words, the aspects are the maximal singularity-free regions *in the workspace*.

The aspects are computed as the connected components of the set obtained by removing the singularity surfaces from the workspace, which can be done easily with the octree model :

$$\cup WA_i = W - S \quad (10)$$

### Application :

For the planar manipulator studied, we get two aspects ($WA_1$ and $WA_2$), where $Det(A)>0$ and $Det(A)<0$, respectively. The singular surface of figure 5 is the common boundary of the two aspects (figure 6).

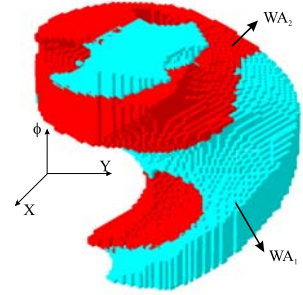

*Figure 6 : Octree model of the aspects*

### 3.2 NON-SINGULAR CONFIGURATION CHANGING TRAJECTORIES

In **[Innocenti 92]**, a non-singular configuration changing trajectory was found for the planar manipulator. However, it appears that this trajectory passes close to a singular configuration. We have been able to confirm that non-singular configuration changing trajectories do exist for this robot. For the following input joint values :

$$\rho_1 = 14.98 \quad \rho_2 = 15.38 \quad \rho_3 = 12.0 \quad (11)$$

The solutions to direct kinematic problem are (Table 1) .

We notice that solutions -2-, -3-, -6- are in the same aspect ($WA_1$) (Figure 7).

This example clearly shows that the aspects are not the uniqueness domains. Additional surfaces have to be defined for separating the solutions.

|   | x | y | φ |
|---|---|---|---|
| 1 | -8.715 | 12.183 | -0.987 |
| 2 | -5.495 | -13.935 | -0.047 |
| 3 | -14.894 | 1.596 | 0.244 |
| 4 | -13.417 | -6.660 | 0.585 |
| 5 | 14.920 | -1.337 | 1.001 |
| 6 | 14.673 | -3.013 | 2.133 |

*Table 1 : Six direct kinematic solutions for the planar manipulator*

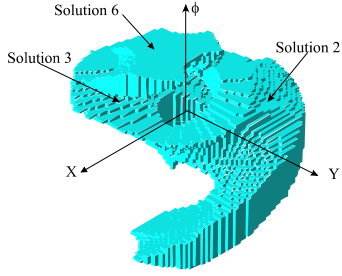

*Figure 7 : Three direct kinematic solutions in one single aspect*

## 4. CHARACTERISTIC SURFACES

### 4.1 DEFINITION

The characteristic surfaces were initially introduced in **[Wenger 92]** for serial manipulators.

This definition is restated here for the case of parallel manipulators.

**Definition 2 :**

Let *WA* an aspect in workspace *W*. The *characteristic surfaces* of the aspect *WA*, denoted $S_c(WA)$, are defined as the preimage in *WA* of the boundary $\overline{WA}$ that delimits *WA* (Figure 8) :

$$S_c(WA) = g^{-1}\left(g\left(\overline{WA}\right)\right) \cap WA \qquad (12)$$

where :
- *g* is defined as in (7)
- $g^{-1}$ is a notation. Let $B \subset Q$ :
$$g^{-1}(B) = \{X \in W \,/\, g(X) \in B\}$$

The boundaries $\overline{WA}$ of *WA* are composed of :
- the type-2 singularities ;
- the type-1 singularities (limits of the actuated joints).

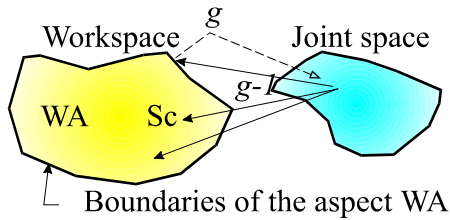

*Figure 8 : Definition of the characteristic surfaces (scheme)*

### 4.2 CASE OF THE PLANAR FULLY PARALLEL MANIPULATOR OF TYPE 3 - RPR

The characteristic surfaces are computed using definition (12). The singular surfaces are scanned and their preimages in the joint space are calculated using *g*. The resulting inverse singularities are mapped back into the workspace using the direct kinematic model. We get one characteristic surfaces for each aspect, denoted $S_{c1}$ and $S_{c2}$, respectively (Figure 9).

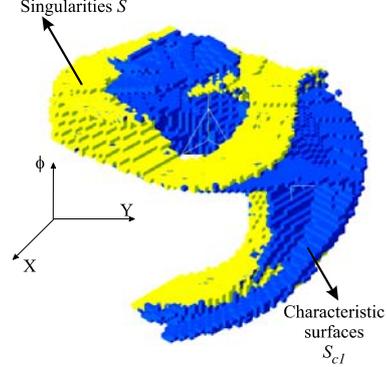

*Figure 9 : Octree model of the singularities and of the characteristic surfaces $S_{c1}$*

## 5. UNIQUENESS DOMAINS

### 5.1 BASIC COMPONENTS AND BASIC DOMAINS

**Definition 3 :**

Let *WA* be an aspect. The *basic regions* of *WA*, denoted $\{WAb_i, i \in I\}$, are defined as the connected components of the set $WA \dotdiv S_C(WA)$ ($\dotdiv$ means the difference between sets). The *basic regions* induce a partition on *WA* :

$$WA = \left(\cup_{i \in I} WAb_i\right) \cup S_C(WA) \qquad (13)$$

**Definition 4 :**

Let $QA_{bi} = g(Ab_i)$, $QA_{bi}$ is a domain in the joint space *Q* called *basic components*. Let *WA* an aspect and *QA* its image under *g*. The following relation holds :

$$QA = \left(\cup_{i \in I} QAb_i\right) \cup g(S_C(WA)) \qquad (14)$$

**Proposition 1 :**

The *basic components* of a given aspect are either coincident, or disjoint sets of *Q*.

Due to space limitation, no proof is provided in this paper. They can be found in **[Chablat 96b]**. The following can be shown :

**Theorem 1 :**

The restriction of *g* to any basic region is a bijection. In other words, there is only one direct solution in each basic region.

**Application :**

The basic regions are calculated as the connected components of the set obtained by removing the characteristic surfaces from the aspects :

$$\cup WA_{bi} = WA_i - S_{ci} \qquad (15)$$

We obtain thus 28 basic regions for the planar manipulator at hand (Figure 10).

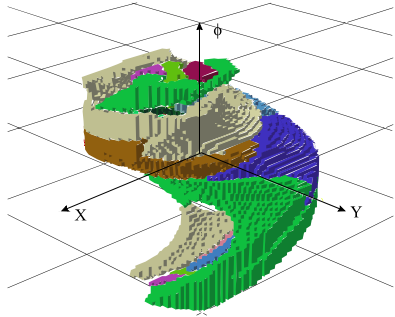

Figure 10 : Octree model of the basic regions

The basic components are computed as :

$$QA_i = g(WAb_i) \qquad (16)$$

We notice that, in accordance with proposition 1, the basic components are either coincident or disjoint. The coincident components yield domains with 2, 4 or 6 solutions for the direct kinematic problem : the upper two domains contain two coincident basic components, the middle four domains are composed of four coincident domains, and the last two domains contain six coincident basic components (Figure 11).

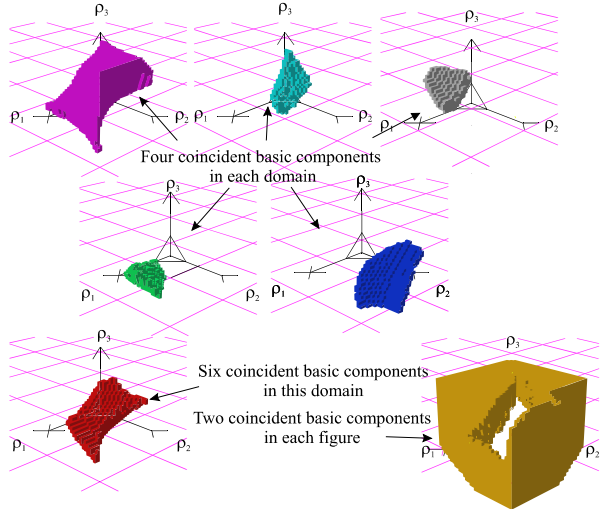

Figure 11 : Octree model of the basic components

## 5.2 UNIQUENESS DOMAINS

Theorem 1 yields sufficient conditions for defining domains of the workspace (the basic regions) where there is one unique solution for the direct kinematic problem. However, the basic regions are not the maximal uniqueness domains. The following theorem 2 intends to define the larger uniqueness domains in the workspace.

**Theorem 2 :**

The uniqueness domains $Wu_k$ are the union of two sets : the set of adjacent basic regions ($\cup_{i \in I'} WAb_i$) of the same aspect $WA$ whose respective preimages are disjoint basic components, and the set $S_c(I')$ of the characteristic surfaces which separate these basic components :

$$Wu_k = \left(\cup_{i \in I'} WAb_i\right) \cup Sc(I') \qquad (17)$$

with $I' \subset I$ such as $\forall i_1, i_2 \in I'$, $g(WAb_{i_1}) \cap g(WAb_{i_2}) = \emptyset$.

Proof of this theorem can be found in **[Chablat 96b].**

**Application :**

To build the uniqueness domains, we have to consider the adjacent basic regions corresponding to disjoint, adjacent basic components.

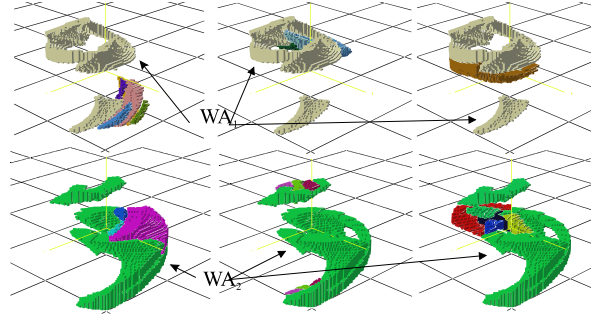

Figure 12 : Octree model of the six uniqueness domains

Six uniqueness domains have been found for the planar manipulator (Figure 12), that is as many as the number of direct kinematic solutions. Note that in general, the number of uniqueness domains should be always more or equal to the maximal number of direct kinematic solutions. We notice that, in the joint space, a non-singular configuration changing trajectory has to go through a domain made of only two coincident basic components, that is, a domain where there are only two direct kinematic solutions (one in each aspect). In addition, it is worth noting that the domain with six coincident basic components map into six basic regions which are linked together by singular surfaces. This means that a non-singular configuration changing trajectory cannot be a straightforward motion between two such basic regions, but should pass through another intermediate basic region (Figure 13).

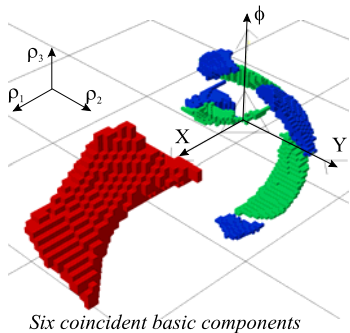

*Six coincident basic components*

*Figure 13 : Basic regions and basic components with 6 solutions to the direct kinematic problem*

## 6. CONCLUSION

The problem of determining the uniqueness domains in the workspace of parallel manipulators has been studied in this paper. The aspects, originally introduced for serial manipulators, have been redefined here as the largest singularity-free regions in the workspace. The aspects were shown to be divided into distinct basic regions where there is only one solution to the direct kinematic problem. These regions are separated by the singular surfaces plus additional surfaces referred to as characteristic surfaces. Physically, the basic regions separate the different solutions to the direct kinematic problem : given a point in the joint space, the corresponding configurations of the moving platform are distributed in the different basic regions. The maximal uniqueness domains have been defined as the union of adjacent basic regions whose preimages in the joint space are not coincident. All results have been illustrated with a 3-DOF planar RPR-parallel manipulator. An octree model of space and specific enrichment techniques have been used for the construction of all sets. This work brings preliminary material to further investigations like trajectory planning **[Merlet 94]**, which is the subject of current research work from the authors.